\RequirePackage{algorithm}
\RequirePackage{algorithmic}

\documentclass[runningheads]{llncs}
\usepackage{graphicx}

\usepackage{tikz}
\usepackage{comment}
\usepackage{amsmath,amssymb} 
\usepackage{color}
\usepackage{microtype}
\usepackage{graphicx}
\usepackage{subfigure}
\usepackage{booktabs}
\usepackage{textcomp}
\usepackage{siunitx}
\usepackage{courier}
\usepackage{adjustbox} 
\usepackage{listings}

\usepackage{float}  
\usepackage[accsupp]{axessibility}  

\makeatletter
\def\blfootnote{\gdef\@thefnmark{}\@footnotetext}
\makeatother

\newcommand{\tr}{\textit{tr}}
\newcommand{\argmin}{\text{argmin }}
\newtheorem{prop}{Proposition}

\begin{document}
\pagestyle{headings}
\mainmatter

\title{Contrasting quadratic assignments for set-based representation learning} 

\titlerunning{Contrasting quadratic assignments for set-based representation learning}

\author{
Artem Moskalev \inst{1} \and
Ivan Sosnovik \inst{1} \and
Volker Fischer \inst{2} \and 
Arnold Smeulders \inst{1}
}
\authorrunning{A. Moskalev et al.}
%
\institute{
UvA-Bosch Delta Lab, University of Amsterdam \\ \email{\{a.moskalev,i.sosnovik,a.w.m.smeulders\}@uva.nl} 
\and
Bosch Center for AI (BCAI) \\
\email{volker.fischer@de.bosch.com}
}

\maketitle

\begin{abstract}

The standard approach to contrastive learning is to maximize the agreement between different views of the data. The views are ordered in pairs, such that they are either positive, encoding different views of the same object, or negative, corresponding to views of different objects. The supervisory signal comes from maximizing the total similarity over positive pairs, while the negative pairs are needed to avoid collapse. In this work, we note that the approach of considering individual pairs cannot account for both intra-set and inter-set similarities when the sets are formed from the views of the data. It thus limits the information content of the supervisory signal available to train representations. We propose to go beyond contrasting individual pairs of objects by focusing on contrasting objects as sets. For this, we use combinatorial quadratic assignment theory designed to evaluate set and graph similarities and derive set-contrastive objective as a regularizer for contrastive learning methods. We conduct experiments and demonstrate that our method improves learned representations for the tasks of metric learning and self-supervised classification.

\blfootnote{Source code: \url{http://github.com/amoskalev/contrasting\_quadratic}}

\end{abstract}

\section{Introduction}

The common approach to contrastive learning is to maximize the agreement between individual views of the data \cite{linsker88self,jaiswal2021survey}. The views are ordered in pairs, such that they either positive, encoding different views of the same object, or negative, corresponding to views of different objects. The pairs are compared against one another by a contrastive loss-objective \cite{oord2019representation,wang2020hypersphere,chen2020simple}. Contrastive learning was successfully applied among others in metric learning \cite{hermans2017defense}, self-supervised classification \cite{chen2020simple,he2019moco}, and pre-training for dense prediction tasks \cite{wang2020DenseCL}. However, when two views of an object are drastically different, the view of object A will resemble the same view of object B much more than it resembles the other view of object A. By comparing individual pairs, the common patterns in the differences between two views will not be exploited. In this paper, we propose to go beyond contrasting individual pairs of objects and focus on contrasting sets of objects.

\begin{figure}[t!]
  \centering
    \includegraphics[width=0.99\linewidth]{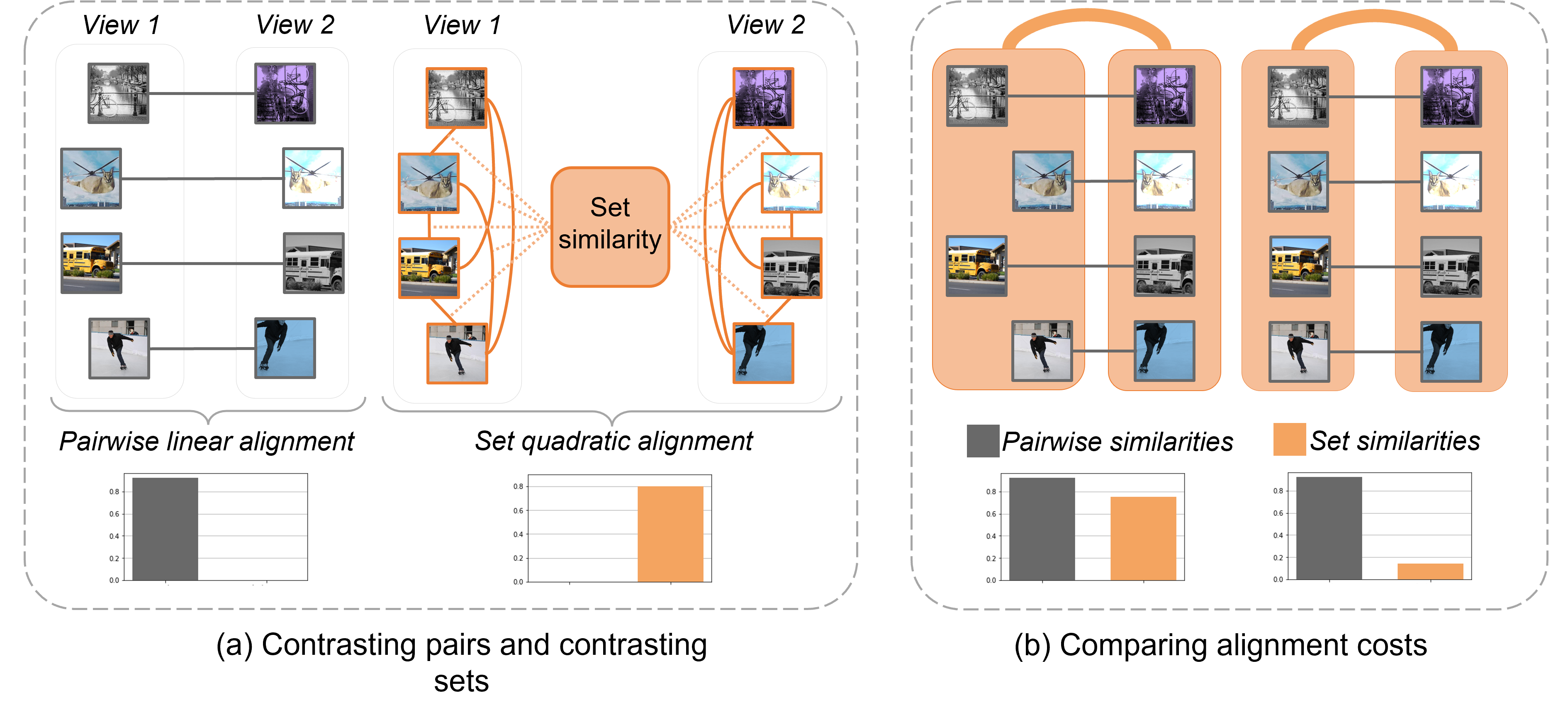}
    \caption{(a) Contrasting views by similarity over pairs (pairwise linear alignment) versus similarity over sets with set-alignment (set-wise quadratic alignment). (b) Comparing total pairwise similarities versus set similarities for different configurations of representation graphs. In both configurations, the total similarity over pairs remains the same, being unable to discriminate between the internal structures of different views as quadratic alignment can.
    }
  \label{fig:intro}
\end{figure}

In contrastive learning, the best alignment of objects follows from maximizing the total similarity over positive pairs, while the negative pairs are needed to encourage diversity. The diversity in representations is there to avoid collapse \cite{tschannen2020On}. We note that one number, the total similarity from contrasting individual pairs, cannot both account for the inter-set similarities of the objects from the same view and the intra-set similarities with objects from another view. Therefore, considering the total similarity over pairs essentially limits the information content of the supervisory signal available to the model. We aim to exploit the information from contrasting sets of objects rather than contrasting objects pairwise only, further illustrated in Figure \ref{fig:intro}a. From the perspective of set assignment theory \cite{burkard2013qap}, two sets may expose the same linear pairwise alignment costs, yet have different internal structures and hence have different set alignment costs, see Figure \ref{fig:intro}b. Therefore, contrastive learning from sets provides a richer supervisory signal. 

To contrast objects as sets, we turn to combinatorial quadratic assignment theory \cite{burkard2013qap} designed to evaluate set and graph similarities. Quadratic assignment has advanced from linear assignment \cite{Burkard98linearassignment}, which relies on pairwise similarities between objects. We pose contrastive learning by structural risk minimization \cite{tsochantaridis05marginSVM} in assignment problems. In this, pairwise contrastive learning methods emerge from the linear assignment case. We aim to extend the contrastive objective to the set level by generalizing the underlying assignment problem from linear to quadratic, as illustrated in Figure \ref{fig:intro}a. Since directly computing set similarities from quadratic alignment is computationally expensive, we derive an efficient approximation. It can be implemented as a regularizer for existing contrastive learning methods. We provide a theory for the proposed method from the perspective of assignment problems and dependence maximization. And, we experimentally demonstrate the advantages of contrasting objects by quadratic alignment.

We make the following contributions: 
\begin{itemize}
    \item Using combinatorial assignment theory, we propose to learn representations by contrasting objects as sets rather than as pairs of individual instances.
    \item We propose a computationally efficient implementation, where the set contrastive part is implemented as a regularizer for existing contrastive learning methods. 
    \item We demonstrate that set-contrastive regularization improves recent contrastive learning methods for the tasks of metric learning and self-supervised classification.
\end{itemize}

As a byproduct of viewing representation learning through the lens of combinatorial assignment theory, we additionally propose SparseCLR contrastive loss, a modification of the popular InfoNCE \cite{oord2019representation} objective. Different from InfoNCE, SparseCLR enjoys sparse support over negative pairs, which permits better use of hard negative examples. Our experiments demonstrate that such an approach improves the quality of learned representations for self-supervised classification.
\section{Related work}
\label{sec:related}

\subsection{Contrastive learning}

In contrastive learning, a model is trained to align representations of the data from different views (modalities), which dates back to the work of Becker \textit{et al.} \cite{Becker1992SelforganizingNN}. Contrastive learning is made up of a joint embedding architecture like Siamese networks \cite{Bromley1993siamese} and a contrastive objective function. The network maps different views of the data into the embedding space, where the alignment between embeddings is measured by contrasting positive and negative pairs. The pairs either from a complete set of observations \cite{peng19comic,chen2020simple,patacchiola2020self,chen2020big} or from partially-complete views \cite{huang2020partially,yang2021partially,lin2021completer,yang2022robust}. And as a contrastive objective, contrastive loss functions are used \cite{oord2019representation,chen2020simple,he2019moco,wang2020hypersphere}. Van den Oord \textit{et al.} \cite{oord2019representation} derived the InfoNCE contrastive loss as a mutual information lower bound between different views of the signal. In computer vision, Chen \textit{et al.} apply the InfoNCE to contrast the images with their distorted version. In PIRL \cite{misra2019selfsupervised}, the authors propose to maintain a memory bank of image representations to improve the generalization of the model by sampling negative samples better. Along the same lines, MoCO \cite{he2019moco,chen2020mocov2} proposes a running queue of negative samples and a momentum-encoder to increase the contrasting effect of negative pairs. All of these contrastive methods are based on measuring the alignment between individual pairs of objects. This approach does not account for patterns in the views of objects beyond contrasting them pairwise. We aim to extend contrastive learning to include set similarities.

\subsection{Information maximization methods}
In contrastive learning information maximization methods aim to improve contrastive representations by maximizing the information content of the embeddings \cite{bardes2022vicreg}. In W-MSE \cite{ermolov2020whitening}, batch representations are passed through a whitening, Karhunen-Loève transformation before computing the contrastive loss. Such transformations help to avoid collapsing representations when the contrastive objective can be minimized without learning the discriminative representations. In \cite{zbontar2021barlow}, the authors follow the redundancy-reduction principle to design a loss function to bring the cross-correlation matrix of representations from different views close to identity. In the recent work of Zbontar \textit{et al.}, the authors extend the above formulation of Barlow Twins with an explicit variance term, which helps to stabilize the training. In this paper, we also seek to maximize the information content of the embeddings but we aim to do so by computing the rich representation of set similarities between data views. In contrast to existing methods, our approach does not require any additional transformations like whitening and can be easily incorporated into other contrastive learning methods.

\subsection{Distillation methods}
Another branch of self-supervised methods is not based on contrastive learning but rather based on knowledge distillation \cite{Hinton2015DistillingTK}. Instead of contrasting positive and negative pairs of samples, these methods try to predict one positive view from another. We discuss them here for completeness. BYOL \cite{grill2020bootstrap} uses a student network that learns to predict the output of a teacher network. SimSiam \cite{chen2020simsiam} simplifies the BYOL by demonstrating that the stop-gradient operation suffices to learn a generalized representation from distillation. In SWaV \cite{caron2020unsupervised}, the authors combine online clustering and distillation by predicting swapped cluster assignments. These self-supervised methods have demonstrated promising results in the task of learning representations. We prefer to follow a different path of contrastive learning, where we do not exclude the possibility that set-based methods may also be applicable to distillation.

\subsection{Assignment theory} Generally, optimal assignment problems can be categorized into linear and higher-order assignments. Linear assignment problems can be viewed as a transportation problem over a bipartite graph, where the transportation distances are defined on individual pairs of nodes. Linear assignment problems have many real-world applications such as scheduling or vehicle routing, we refer to \cite{Burkard98linearassignment}. Higher-order or in particular Quadratic assignment problems \cite{burkard87qap,burkard2013qap} extend the domain of the transportation distances from individual pairs to sets of objects.  Where the linear assignment problem seeks to optimally match individual instances, its quadratic counterpart matches the inter-connected graphs of instances, bringing additional structure to the task. In this work, we aim to exploit linear and quadratic assignment theory to rethink contrastive learning. 

\subsection{Structured predictions}
In a structured prediction problem, the goal is to learn a structured object such as a sequence, a tree, or a permutation matrix \cite{bakir2007structured}. Training a structure is non-trivial as one has to compute the loss on the manifold defined by the structured output space. In the seminal work of Tsochantaridis \textit{et al.} \cite{tsochantaridis05marginSVM}, the authors derive a structured loss for support vector machines and apply it to a sequence labeling task. Later, structured prediction was applied to learn parameters of constraint optimization problems \cite{Hazan2010dlm,blondel2020learning}. In this work, we utilize structured prediction principles to implement set similarities into contrastive losses.

\section{Background}
\label{sec:background}

We start by formally introducing contrastive learning and linear assignment problems. Then, we argue about the connection between these two problems and demonstrate how one leads to another. This connection is essential for the derivation of our contrastive method.

\subsection{Contrastive representation learning} Consider a dataset $\mathcal{D}=\{ d_i \}_{i=1}^{N}$ and an encoder function $f_{\theta}: \mathcal{D} \longrightarrow \mathbb{R}^{N \times E}$, which outputs an embedding vector of dimension $E$ for each of the objects in $\mathcal{D}$. The task is to learn a such $f_{\theta}$ that embeddings of objects belonging to the same category are pulled together, while the embeddings of objects from different categories are pushed apart.

As category labels are not available during training, the common strategy is to maximize the agreement between different views of the instances, i.e. to maximize the mutual information between different data modalities. The mutual information between two latent variables $X, Y$, can be expressed as:

\begin{equation}
    \label{eq:mi}
    MI(X,Y) = H(X) - H(X | Y)
\end{equation}

where $X$ and $Y$ correspond to representations of two different views of a dataset. Minimizing the conditional entropy $H(X | Y)$ aims at reducing uncertainty in the representations from one view given the representations of another, while $H(X)$ enforces the diversity in representations and prevents trivial solutions. In practice, sample-based estimators of the mutual information such as InfoNCE \cite{oord2019representation} or NT-Xent \cite{chen2020simple} are used to maximize the objective in \eqref{eq:mi}.

\subsection{Linear assignment problem}

Given two input sets, $\mathcal{A} = \{ a_{i} \}_{i=1}^{N}$ and $\mathcal{B} = \{ b_{j} \}_{j=1}^{N}$, we define the inter-set similarity matrix $\textbf{S} \in \mathbb{R}^{N \times N}$ between each element in set $\mathcal{A}$ and each element in $\mathcal{B}$. This similarity matrix encodes \textit{pairwise} distances between the elements of the sets, \textit{i.e.} $[\textbf{S}]_{i,j} = \phi(a_{i}, b_{j})$, where $\phi$ is a distance metric. The goal of the linear assignment problem is to find a one-to-one assignment $\hat{y}(\textbf{S})$, such that the sum of distances between assigned elements is optimal. Formally:

\begin{equation}
    \label{eq:lap}
    \hat{y}(\textbf{S}) = \underset{\textbf{Y} \in \Pi} {\argmin} \tr(\textbf{S}\textbf{Y}^T)
\end{equation}

where $\Pi$ corresponds to a set of all $N \times N$ permutation matrices.

The linear assignment problem in \eqref{eq:lap} is also known as the bipartite graph matching problem. It can be efficiently solved by linear programming algorithms \cite{bertsimas98intro}.

\subsection{Learning to produce correct assignments}
\label{sec:learning_lap}

Consider two sets, $\mathcal{D}_{\mathcal{Z}_1}$ and $\mathcal{D}_{\mathcal{Z}_2}$, which encode two different views of the dataset $\mathcal{D}$ under two different views $\mathcal{Z}_1$ and $\mathcal{Z}_2$. By design, the two sets consist of the same instances, but the modalities and the order of the objects may differ. With the encoder, which maximizes the mutual information, objects in both sets can be uniquely associated with one another by comparing their representations as the uncertainty term in \eqref{eq:mi} should be minimized. In other words, $\hat{y}(\textbf{S}) = \textbf{Y}_{gt}$, where $\textbf{Y}_{gt} \in \Pi$ is a ground truth assignment between elements of $\mathcal{D}_{\mathcal{Z}_1}$ and $\mathcal{D}_{\mathcal{Z}_2}$.

Thus, a natural objective to supervise an encoder function is to train it to produce the correct assignment between the different views of the data. As the assignment emerges as a result of the optimization problem, we employ a structured prediction framework \cite{tsochantaridis05marginSVM} to define a structural loss from the linear assignment problem as follows:

\begin{equation}
    \label{eq:salo}
    L(\textbf{S}, \textbf{Y}_{gt}) = \tr (\textbf{S} \textbf{Y}_{gt}^T) - \underset{\textbf{Y} \in \Pi} {\min} \tr(\textbf{S}\textbf{Y}^T)
\end{equation}

where $\textbf{S}=\phi \big{(} f_{\theta} (\mathcal{D}_{\mathcal{Z}_1}), f_{\theta} (\mathcal{D}_{\mathcal{Z}_2}) \big{)}$. Note that $L \geq 0$ and $L = 0$ only when the similarities produced by an encoder lead to the correct optimal assignment. Intuitively, the structured linear\footnote{We emphasize that the term \textit{linear} here is only used with regard to an underlying assignment problem.} assignment loss in \eqref{eq:salo} encodes the discrepancy between the true cost of the assignment and the cost induced by an encoder.

By minimizing the objective in \eqref{eq:salo}, we train the encoder $f_{\theta}$ to correctly assign objects from one view to another. In practice, it is desirable that such assignment is robust under small perturbations in the input data. The straightforward way to do so is to enforce a separation margin $m$ in the input similarities as $\textbf{S}_m = \textbf{S} + m\textbf{Y}_{gt}$. Then, the structured linear assignment loss with separation margin reduces to a known margin Triplet loss \cite{hermans2017defense}.

\begin{prop}
\label{pr:prop1}
The structured linear assignment loss $L(\textbf{S}_m, \textbf{Y}_{gt})$ with separation margin $m \geq 0$ is equivalent to the margin triplet loss.
\end{prop} 

\paragraph{Mining strategies.} By default, the loss in \eqref{eq:salo} enforces a one-to-one negative pair mining strategy due to the structural domain constraint $\textbf{Y} \in \Pi$. By relaxing this domain constraint to row-stochastic binary matrices, we arrive at the known batch-hard mining \cite{hermans2017defense}. This is essential to have a computationally tractable implementation of structured assignment losses.

\paragraph{Smoothness.} An immediate issue when directly optimizing the structured linear assignment loss is the non-smoothness of the minimum function in \eqref{eq:salo}. It is known that optimizing smoothed functions can be more efficient than directly optimizing their non-smooth counterparts \cite{beck12smoothingand}. The common way to smooth a minimum is by \textit{log-sum-exp} approximation. Thus, we can obtain a smoothed version of structured linear assignment loss:

\begin{equation}
    \label{eq:smooth-salo}
    L_{\tau}(\textbf{S}, \textbf{Y}_{gt}) = \tr(\textbf{S} \textbf{Y}_{gt}^T) + \tau \log \sum_{\textbf{Y} \in \Pi} \exp(-\frac{1}{\tau} \tr(\textbf{S}\textbf{Y}^T) )
\end{equation}

where $\tau$ is a temperature parameter controlling the degree of smoothness. Practically, the formulation in \eqref{eq:smooth-salo} requires summing $N!$ terms, which makes it computationally intractable under the default structural constraints $\textbf{Y} \in \Pi$. Fortunately, similar to the non-smooth case, we can utilize batch-hard mining, which leads to $O(N^2)$ computational complexity. The smoothed structured linear assignment loss with batch-hard mining reduces to known normalized-temperature cross entropy \cite{chen2020simple} also known as the InfoNCE \cite{oord2019representation} loss.

\begin{prop}
\label{pr:prop2}
The smoothed structured linear assignment loss $L_{\tau}({\textbf{S}}, \textbf{Y}_{gt})$ with batch-hard mining is equivalent to the normalized-temperature cross entropy loss. 
\end{prop}

\paragraph{Connection to mutual information} It is known that the InfoNCE objective is a lower bound on the mutual information between the representations of data modalities \cite{oord2019representation}. Thus, Propositions \ref{pr:prop1} and \ref{pr:prop2} reveal a connection between mutual information maximization and minimization of structured losses for assignmnet problems. In fact, the assignment cost $\tr(\textbf{S} \textbf{Y}_{gt}^T)$ in \eqref{eq:lap} is related to the conditional entropy $H(X | Y)$, while $\min_{\textbf{Y} \in \Pi} \tr(\textbf{S}\textbf{Y}^T)$ aims to maximize the diversity in representations. This connection allows for consideration of contrastive representation learning methods based on InfoNCE as a special case the structured linear assignment loss.

\section{Extending contrastive losses}
\label{sec:method2}

We next demonstrate how to exploit the connection between contrastive learning and assignment problems to extend contrastive losses the set level. As a byproduct of this connection, we also derive the SparseCLR contrastive objective.

\subsection{Contrastive learning with quadratic assignments on sets}
\label{sec:ho_method}

\subsubsection{Quadratic assignment problem}
As in the linear case, we are given two input sets $\mathcal{A}, \mathcal{B}$ and the inter-set similarity matrix $\textbf{S}$. For the Quadratic Assignment Problem (QAP), we define intra-set similarity matrices $\textbf{S}_{\mathcal{A}}$ and $\textbf{S}_{\mathcal{B}}$ measuring similarities within the sets $\mathcal{A}$ and $\mathcal{B}$ respectively, \textit{i.e.} $[\textbf{S}_{\mathcal{A}}]_{ij} = \phi(a_i, a_j)$. A goal of the quadratic assignment problem is to find a \textit{one-to-one} assignment $\hat{y}_{\mathcal{Q}}(\textbf{S}, \textbf{S}_{\mathcal{A}}, \textbf{S}_{\mathcal{B}}) \in \Pi$ that maximizes the set similarity between $\mathcal{A}, \mathcal{B}$, where the set similarity is defined as follows:

\begin{equation}
\begin{split}
    \label{eq:qap}
    \mathcal{Q}(\textbf{S}, \textbf{S}_{\mathcal{A}}, \textbf{S}_{\mathcal{B}}) =
    \underset{\textbf{Y} \in \Pi} {\min} \big{\{} \tr(\textbf{S}\textbf{Y}^T) + \tr(\textbf{S}_{\mathcal{A}} \textbf{Y} \textbf{S}_{\mathcal{B}}^{T} \textbf{Y}^{T}) \big{\}}
\end{split}
\end{equation}

Compared to the linear assignment problem, the quadratic term $\mathcal{Q}(\textbf{S}, \textbf{S}_{\mathcal{A}}, \textbf{S}_{\mathcal{B}})$ in \eqref{eq:qap} additionally measures the discrepancy in internal structures between sets $\textbf{S}_{\mathcal{A}}$ and $\textbf{S}_{\mathcal{B}}$.

\subsubsection{Learning with quadratic assignments}

Following the similar steps as for the linear assignment problem in Section \ref{sec:learning_lap}, we next define the structured quadratic assignment loss by extending the linear assignment problem in \eqref{eq:salo} with the quadratic term:

\begin{equation}
    \label{eq:ho_loss}
    L_{QAP} = \tr(\textbf{S}\textbf{Y}_{gt}^T) - \mathcal{Q}(\textbf{S}, \textbf{S}_{\mathcal{A}}, \textbf{S}_{\mathcal{B}})
\end{equation}

Minimization of the structured quadratic assignment loss in \eqref{eq:ho_loss} encourages the encoder to learn representations resulting in the same solutions of the linear and quadratic assignment problems, which is only possible when the inter-set and intra-set similarities are sufficiently close \cite{burkard2013qap}. Note that we do not use a ground truth quadratic assignment in $L_{QAP}$, but a ground truth linear assignment objective is used for the supervision. This is due to a quadratic nature of $\mathcal{Q}$, where minimizing the discrepancy between ground truth and predicted assignments is a subtle optimization objective, e.g. for an equidistant set of points the costs of all quadratic assignments are identical.

To compute $L_{QAP}$, we first need to evaluate the quadratic term $\mathcal{Q}(\textbf{S}, \textbf{S}_{\mathcal{A}}, \textbf{S}_{\mathcal{B}})$, which requires solving the quadratic assignment. This problem is known to be notoriously hard to solve exactly even when an input dimensionality is moderate \cite{burkard2013qap}. To alleviate this, we use a computationally tractable upper-bound: 

\begin{equation}
\begin{split}
    \label{eq:ho_aprx}
    L_{QAP} & \leq \tr(\textbf{S}\textbf{Y}_{gt}^T) - \underset{\textbf{Y} \in \Pi} {\min} \tr(\textbf{S} \textbf{Y}^T)
    - \underset{\textbf{Y} \in \Pi} {\min} \tr(\textbf{S}_{\mathcal{A}} \textbf{Y} \textbf{S}_{\mathcal{B}}^T \textbf{Y}^T) \\
    & \leq L(\textbf{S}, \textbf{Y}_{gt}) - \langle \lambda_{\mathcal{A}}, \lambda_{\mathcal{B}} \rangle_{-}
\end{split}
\end{equation}

where $\langle \lambda_{\mathcal{A}}, \lambda_{\mathcal{B}} \rangle_{-}$ corresponds to a minimum dot product between eigenvalues of matrices $S_{A}$ and $S_{B}$. 

The first inequality is derived from Jensen's inequality for \eqref{eq:ho_loss} and the second inequality holds due to the fact that $\langle \lambda_{F}, \lambda_{D} \rangle_{-} \leq \tr(\textbf{F} \textbf{X} \textbf{D}^T \textbf{X}^T) \leq \langle \lambda_{F}, \lambda_{D} \rangle_{+}$ for symmetric matrices $\textbf{F}, \textbf{D}$ and $\textbf{X} \in \Pi$ as shown in \textit{Theorem 5} by Burkard \cite{burkard2013qap}. The above derivations are for the case when the similarity metric is a valid distance function, i.e. minimizing a distances leads to maximizing a similarity. The derivation for the reverse case can be done analogously by replacing \textit{min} with \textit{max} in the optimal assignment problem formulation (we provide more details in supplementary material).

Optimizing the upper-bound in \eqref{eq:ho_aprx} is computationally tractable compared to optimizing the exact version of $L_{QAP}$. Another advantage is that the upper-bound in \eqref{eq:ho_aprx} breaks down towards minimizing the sum of the structured linear assignment loss $L(\textbf{S}, \textbf{Y}_{gt})$ and the regularizing term, which accounts for the set similarity. This allows to easily modify existing contrastive learning approaches like those in \cite{chen2020simple,hermans2017defense,mikolov2013efficient,oord2019representation} that are based on pairwise similarities and thus stem from the linear assignment problem by simply plugging in the regularizing term. We provide a simple pseudocode example demonstrating InfoNCE with the quadratic assignment regularization (Algorithm \ref{alg:infonce_qare}).

\begin{algorithm}[t]
   \caption{Pseudocode for set-based InfoNCE with Euclidean distance similarity and Quadratic Assignment Regularization (QARe).}
   \label{alg:infonce_qare}
   
    \definecolor{codeblue}{rgb}{0.25,0.5,0.5}
    \lstset{
      basicstyle=\fontsize{7.2pt}{7.2pt}\ttfamily\bfseries,
      commentstyle=\fontsize{7.2pt}{7.2pt}\color{codeblue},
      keywordstyle=\fontsize{7.2pt}{7.2pt},
    }
\begin{lstlisting}[language=python]
# f: encoder network
# alpha: weighting for the pairwise contrastive part (linear assignment)
# beta: weighting for the set contrastive part
# N: batch size
# E: dimensionality of the embeddings

for x in loader: # load a batch with N samples
    # two randomly augmented views of x
    y_a, y_b = augment(x)
    
    # compute embeddings
    z_a = f(y_a) # NxE
    z_b = f(y_b) # NxE
    
    # compute inter-set and intra-set similarities
    S_AB = similarity(z_a, z_b) # NxN
    S_A  = similarity(z_a, z_a) # NxN
    S_B  = similarity(z_b, z_b) # NxN
    
    # compute pairwise contrastive InfoNCE loss
    pairwise_term = infonce(S_AB)
    
    # compute eigenvalues
    eigs_a = eigenvalues(S_A) #N
    eigs_b = eigenvalues(S_B) #N
    
    # compute QARe from minimum dot product of eigenvalues
    eigs_a_sorted = sort(eigs_a, descending = True) #N
    eigs_b_sorted = sort(eigs_b, descending = False) #N
    qare = -1*(eigs_a_sorted.T@eigs_b_sorted)
    
    # combine pairwise contrastive loss with QARe
    loss = alpha*pairwise_term + beta*qare/(N^2)
    
    # optimization step
    loss.backward()
    optimizer.step()
\end{lstlisting}
\end{algorithm}

\subsubsection{Computational complexity} Computing the upper-bound in \eqref{eq:ho_aprx} has a computational complexity of $O(N^3)$ as one needs to compute eigenvalues of the intra-set similarity matrices. This is opposed to \eqref{eq:ho_loss} that requires directly computing quadratic assignments, for which it is known there exist no polynomial-time algorithms \cite{burkard2013qap}. The computational complexity can be pushed further down to $O(k^2N)$ by evaluating the only top-k eigenvalues instead of computing all eigenvalues. In supplementary materials, we also provide an empirical analysis of how the proposed approach influences a training time of a baseline contrastive method.

\subsection{SparseCLR}
\label{sec:sparsemax}

In Section \ref{sec:background} we noted that the smoothness of a loss function is a desirable property for optimization and that the \textit{log-sum-exp} smoothing of the structured linear assignment loss yields the normalized temperature cross-entropy objective as a special case. Such an approach, however, is known to have limitations. Specifically, the \textit{log-sum-exp} smoothing yields dense support over samples, being unable to completely rule out irrelevant examples \cite{Niculae2018SparseMAPDS}. That can be statistically and computationally overwhelming and can distract the model from fully utilizing information from hard negative examples.

We propose to use \textit{sparsemax} instead of \textit{log-sum-exp} approximation to alleviate the non-smoothness of the structured linear assignment objective, yet keep the sparse support of the loss function. Here \textit{sparsemax} is defined as the minimum distance Euclidean projection to a k-dimensional simplex as in \cite{Niculae2018SparseMAPDS,Martins2016sparsemax}.

Let $\Tilde{x}: \in \mathbb{R}^{N!}$ be a vector that consists of the realizations of all possible assignment costs for the similarity matrix \textbf{S}. With this we can define SparseCLR as follows:

\begin{equation}
    \label{eq:sparseclr}
    L_{\textit{sparse}}(\textbf{S}, \textbf{Y}_{gt}) = \tr(\textbf{S} \textbf{Y}_{gt}^T) - \frac{1}{2} \sum_{j \in \Omega(\Tilde{x})} \big{(} \Tilde{x}_{j}^{2} - \mathcal{T}^{2}(-\Tilde{x}) \big{)}
\end{equation}

where $\Omega(X) = \{ j \in X : \textit{ sparsemax}(X)_{j} > 0\}$ is the support of \textit{sparsemax} function, and $\mathcal{T}$ denotes the thresholding operator \cite{Martins2016sparsemax}. In practice, to avoid summing over a factorial number of terms, as well as for other methods, we use batch-hard mining strategy resulting in $O(N^2)$ computational complexity for SparseCLR.
\section{Experiments}

In this section, we evaluate the quality of representations trained with and without our Quadratic Assignment Regularization (QARe). We also evaluate the performance of SparseCLR and compare it with other contrastive losses. As we consider the representation learning from the perspective of the assignment theory, we firstly conduct the instance matching experiment, where the goal is to learn to predict the correct assignment between different views of the dataset. Then, we test the proposed method on the task of self-supervised classification and compare it with other contrastive learning approaches. Next, we present ablation studies to visualize the role of the weighting term, when combining QARe with the baseline contrastive learning method.

\subsection{Matching instances from different views}

In this experiment, the goal is to train representations of objects from different views, such that the learned representations provide a correct matching between identities in the dataset. This problem is closely related to a metric learning and can be solved by contrasting views of the data \cite{hermans2017defense}.

\subsubsection{Data} We adopt the CUHK-03 dataset \cite{li2014deepreid} designed for the ReID \cite{hermans2017defense} task. CUHK-03 consist of 1467 different identities recorded each from front and back views. To train and test the models, we randomly divide the dataset into \textit{70/15/15} train/test/validation splits.

\subsubsection{Evaluation} We evaluate the quality of representations learned with contrastive losses by computing the matching accuracy between front and back views. In practice, we first obtain the embeddings of the views of the instances from the encoder and then compute their inter-view similarity matrix using the Euclidean distance. Given this, the matching accuracy is defined as the mean Hamming distance between the optimal assignment from the inter-view similarity matrix and the ground truth assignment. We report the average matching accuracy and the standard deviation over 3 runs with the same fixed random seeds. As a baseline, we chose Triplet with batch-hard mining \cite{hermans2017defense}, InfoNCE \cite{oord2019representation} and NTLogistic \cite{mikolov2013efficient} contrastive losses, which we extend with the proposed QARe.

\begin{table}[t]
\centering
\begin{tabular}{@{}ccccc@{}}
\toprule

\hspace{3mm} \textbf{Method} \hspace{3mm} & 
\hspace{3mm} Triplet \hspace{3mm} & 
\hspace{3mm} SparseCLR \hspace{3mm} & 
\hspace{3mm} InfoNCE \hspace{3mm} & 
\hspace{3mm} NTLogistic \hspace{3mm} \\ \midrule

\textit{pairwise} & 
54.85$\pm$\small{0.77} & 
53.03$\pm$\small{0.43} & 
57.87$\pm$\small{1.19} & 
40.30$\pm$\small{0.42} \\

\textit{+QARe} & 
\textbf{58.48$\pm$\small{1.67}} & 
\textbf{54.84$\pm$\small{1.40}} & 
\textbf{61.96$\pm$\small{1.38}} & 
\textbf{43.48$\pm$\small{1.41}} \\ 
\bottomrule
\end{tabular}
\vspace{1mm}
\caption{Matching accuracy for instances from different views of CUHK-03 dataset. The \textit{pairwise} corresponds to the contrastive losses acting on the level of pairs. \textit{+QARe} denotes methods with quadratic assignment regularization. Best results are in bold.}
\label{tab:matching}
\end{table}

\subsubsection{Implementation details}
As the encoder, we use ResNet-18 \cite{he2016deep} with the top classification layer replaced by the linear projection head that outputs feature vectors of dimension 64. To obtain representations, we normalize the feature vectors to be L2 unit-norm. 

We train the models for 50 epochs using Adam optimizer \cite{kingma2014adam} with the cosine annealing without restarts \cite{loshchilov2016sgdr}. Initial learning rate is 0.01 and a batch size is set to 128. During the training, we apply random color jittering and horizontal flip augmentations. To compute the test matching accuracy, we select the best model over the epochs based on the validation split. We provide detailed hyperparameters of the losses, QARe weighting, and augmentations in the supplementary material.

\subsubsection{Results}
The results are reported in Table \ref{tab:matching}. As can be seen, adding QARe regularization to a baseline contrastive loss leads to a better matching accuracy, which indicates an improved quality of representations. Notably, QARe delivers 3.6\% improvement when combined with Triplet loss and 4.1\% increase in accuracy with the InfoNCE objective. This demonstrates that the quadratic assignment regularization on sets helps the model to learn generalized representation space.

\subsection{Self-supervised classification}

Next, we evaluate the quadratic assignments on sets in the task of self-supervised classification. The goal in this experiment is to learn a representation space from an unlabeled data that provides a linear separation between classes in a dataset. This is a common problem to test contrastive learning methods.

\subsubsection{Evaluation} We follow the standard linear probing protocol \cite{kolesnikov2019revisiting}. The evaluation is performed by retrieving embeddings from a contrastively trained backbone and fitting a linear classifier on top of the embeddings. Since linear classifiers have a low discriminative power on their own, they heavily rely on input representations. Thus, a higher classification accuracy indicates a better quality of learned representations. As the baseline for pairwise contrastive learning methods, we select popular SimCLR and proposed SparseCLR. We extend the methods to the set level by adding the quadratic assignment regularization. We compare the set-level contrastive methods with other self-supervised \cite{caron2018deep,gidaris2018unsupervised} and contrastive approaches \cite{chen2020simple,caron2020unsupervised}.

\begin{table*}[t!]
 \begin{adjustbox}{width=1\columnwidth,center}
  \begin{tabular}{lcccccc}
    \toprule
     & \multicolumn{3}{c}{\textbf{Conv-4}} & \multicolumn{3}{c}{\textbf{ResNet-32}} \\
    \cmidrule[0.1pt](r){2-4} \cmidrule[0.1pt](l){5-7}
    \textbf{Method} &
    \textbf{CIFAR-10} & \textbf{CIFAR-100} & \textbf{tiny-ImageNet} &
    \textbf{CIFAR-10} & \textbf{CIFAR-100} & \textbf{tiny-ImageNet} \\
    \midrule
    Supervised \text{(upper bound)} & 
    80.46$\pm$\small{0.39} & 49.29$\pm$\small{0.85}  & 36.47$\pm$\small{0.36} &
    90.87$\pm$\small{0.41} & 65.32$\pm$\small{0.22}  & 50.09$\pm$\small{0.32} \\
    Random Weights \text{(lower bound)} & 
    32.92$\pm$\small{1.88} & 10.79$\pm$\small{0.59}  & 6.19$\pm$\small{0.13} &
    27.47$\pm$\small{0.83} &  7.65$\pm$\small{0.44}  & 3.24$\pm$\small{0.43} \\
    \cmidrule(l){1-7}
    DeepCluster \cite{caron2018deep} & 
    42.88$\pm$\small{0.21} & 21.03$\pm$\small{1.56}  & 12.60$\pm$\small{1.23} &
    43.31$\pm$\small{0.62} & 20.44$\pm$\small{0.80}  & 11.64$\pm$\small{0.21} \\
    RotationNet \cite{gidaris2018unsupervised} & 
    56.73$\pm$\small{1.71} & 27.45$\pm$\small{0.80}  & 18.40$\pm$\small{0.95} &
    62.00$\pm$\small{0.79} & 29.02$\pm$\small{0.18}  & 14.73$\pm$\small{0.48} \\
    Deep InfoMax \cite{hjelm2018learning} & 
    44.60$\pm$\small{0.27} & 22.74$\pm$\small{0.21}  & 14.19$\pm$\small{0.13} &
    47.13$\pm$\small{0.45} & 24.07$\pm$\small{0.05}  & 17.51$\pm$\small{0.15} \\
    
    SimCLR$^*$ \cite{chen2020simple} & 
    59.64$\pm$\small{0.93} & 
    30.75$\pm$\small{0.23} & 
    21.15$\pm$\small{0.16} &
    76.45$\pm$\small{0.10} & 
    40.18$\pm$\small{0.13} & 
    23.44$\pm$\small{0.27} 
    \\
    \cmidrule(l){1-7}
    SimCLR+QARe \text{(Ours)} & \textbf{61.54}$\pm$\small{\textbf{1.17}} & \textbf{31.18}$\pm$\small{\textbf{0.41}} & \textbf{21.33}$\pm$\small{\textbf{0.18}} &
    \textbf{76.85}$\pm$\small{\textbf{0.39}} & \textbf{41.20}$\pm$\small{\textbf{0.34}} & \textbf{25.05}$\pm$\small{\textbf{0.18}} 
    \\
    \bottomrule
  \end{tabular}
 \end{adjustbox}
 \centering
 \vspace{1mm}
 \caption{Self-supervised training and linear probing for a self-supervised classification. Average classification accuracy (percentage) and standard deviation over 3 runs with common fixed random seeds are reported. $^*$ we report the results from our reimplementation of SimCLR.}
\label{tab:qare}
\end{table*}

\subsubsection{Implementation details}

We train shallow convolutional (Conv-4) and deep (ResNet-32) backbone encoders to obtain representations. We follow the standard approach and attach an MLP projection head that outputs feature vectors of dimension 64 to compute a contrastive loss. We use the cosine distance to get similarities between embeddings. We provide more details on the encoder architectures in the supplementary material.

We train the models for 200 epochs using Adam optimizer with a learning rate of 0.001 and a batch size is set to 128. For contrastive learning methods, we adopt a set of augmentations from \cite{chen2020simple} to form different views of the data. For linear probing, we freeze the encoder weights and train a logistic regression on top of the learned representation for 100 epochs with Adam optimizer, using a learning rate of 0.001 and a batch size of 128. We provide detailed hyperparameters of the contrastive losses and augmentations in the supplementary material.

We perform training and testing on standard CIFAR-10, CIFAR-100, and tiny-ImageNet datasets. For each dataset, we follow the same training procedure and fine-tune the optimal weighting of the quadratic assignment regularizer.

\begin{figure}[t!]
  \centering
    \includegraphics[width=0.99\linewidth]{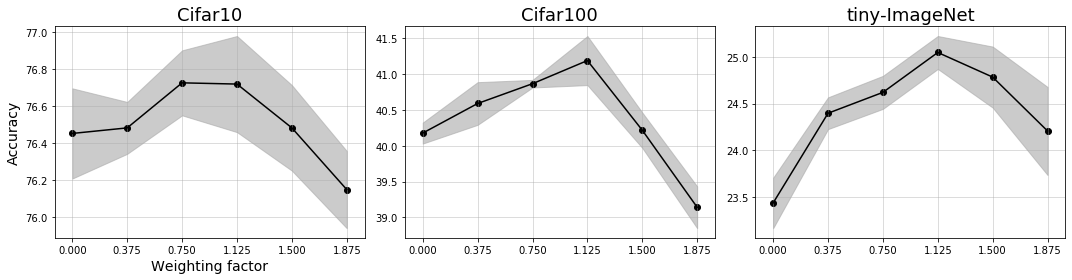}
    \caption{Influence of the weighting of the QARe term in the self-supervised classification for ResNet-32 trained with SimCLR. The accuracy is recorded over 3 runs with common fixed random seeds.}
  \label{fig:ablation}
\end{figure}

\begin{table*}[t]
 \begin{adjustbox}{width=1\columnwidth,center}
  \begin{tabular}{lcccccc}
    \toprule
     & \multicolumn{3}{c}{\textbf{Conv-4}} & \multicolumn{3}{c}{\textbf{ResNet-32}} \\
    \cmidrule[0.1pt](r){2-4} \cmidrule[0.1pt](l){5-7}
    \textbf{Method} &
    \textbf{CIFAR-10} & \textbf{CIFAR-100} & \textbf{tiny-ImageNet} &
    \textbf{CIFAR-10} & \textbf{CIFAR-100} & \textbf{tiny-ImageNet} \\
    \midrule

    SimCLR$^*$ \cite{chen2020simple} & 
    59.64$\pm$\small{0.93} & 
    30.75$\pm$\small{0.23} & 
    21.15$\pm$\small{0.16} &
    \textbf{76.45$\pm$\small{0.10}} & 
    40.18$\pm$\small{0.13} & 
    23.44$\pm$\small{0.27} 
    \\
    SparseCLR (Ours) & 
    59.86$\pm$\small{0.56} & 
    32.04$\pm$\small{0.26} & 
    \textbf{22.41$\pm$\small{0.39}} &
    70.37$\pm$\small{0.12} & 
    41.05$\pm$\small{0.20} & 
    25.87$\pm$\small{0.59} 
    \\
    \cmidrule(l){1-7}
    SparseCLR+QARe (Ours) & 
    \textbf{61.06$\pm$\small{0.39}} & 
    \textbf{33.05$\pm$\small{0.32}} & 
    \textbf{22.45$\pm$\small{0.30}} &
    71.39$\pm$\small{0.30} & 
    \textbf{42.07$\pm$\small{0.35}} & 
    \textbf{27.03$\pm$\small{0.40}} 
    \\
    \bottomrule
  \end{tabular}
 \end{adjustbox}
 \centering
 \vspace{1mm}
 \caption{Self-supervised training on unlabeled data and linear evaluation on labeled data. Comparing SimCLR \cite{chen2020simple} with the proposed SparseCLR and SparseCLR with QARe. Average accuracy and standard deviation over three runs over 3 runs with common fixed random seeds.}
\label{tab:sparse}
\end{table*}

\subsubsection{QARe results}
The results for the quadratic assignment regularization are reported in Table \ref{tab:qare} for SimCLR and in Table \ref{tab:sparse} for proposed SparseCLR. For SimCLR, adding QARe delivers 1.9\% accuracy improvement on CIFAR-10 for the shallow encoder network and up to 1.6 \% improvement for ResNet-32 on tiny-ImageNet. We observed a consistent improvement in other dataset-architecture setups, except tiny-ImageNet with shallow Conv-4 encoder, where the performance gain from adding QARe is modest. For this case, we investigated the training behavior of the models and observed that both SimCLR and SimCLR+QARe for Conv-4 architecture very quickly saturate to a saddle point, where the quality of representations stops improving. Since this does not happen with ResNet-32 architecture, we attribute this phenomenon to the limited discriminative power of the shallow Conv-4 backbone, which can not be extended by regularizing the loss.

For SparseCLR, we observed the same overall pattern. As can be seen from Table \ref{tab:sparse}, extending SparseCLR to set level with QARe delivers 1.6\% accuracy improvement for ResNet-32 on tiny-ImageNet and also steadily improve the performance on other datasets. This indicates that QARe helps to provide a richer supervisory signal for the model to train representations.

\subsubsection{SparseCLR results}
Next, we compare SimCLR against the proposed SparseCLR. As can be seen in Table \ref{tab:sparse}, SparseCLR consistently improves over the baseline on CIFAR-100 and tiny-ImageNet datasets, where it delivers 2.4\% improvement. Surprisingly, we noticed a significant drop in performance for ResNet-32 on CIFAR-10. For this case, we investigated the training behavior of the model and observed that in the case of CIFAR-10, the batch often includes many false negative examples, which results in uninformative negative pairs. Since SparseCLR assigns the probability mass only to a few hardest negative pairs, a lot of false-negative examples in the batch impede obtaining a clear supervisory signal. We assume the problem can be alleviated by using false-negative cancellation techniques \cite{huynh2021fcn}.

\subsection{Ablation study}

Here we illustrate how the weighting of the QARe term influences the quality of representations learned with SimCLR under a linear probing evaluation. We search for the optimal weighting in the range from 0 to 1.875 with a step of 0.125. The results are depicted in Figure \ref{fig:ablation}. In practice, we observed that QARe is not too sensitive to a weighting and the values in the range 0.75-1.25 provide consistent improvement.
\section{Discussion}

In this work, we present set contrastive learning method based on quadratic assignments. Different from other contrastive learning approaches, our method works on the level of set similarities as opposed to only pairwise similarities, which allows improving the information content of the supervisory signal. For derivation, we view contrastive learning through the lens of combinatorial assignment theory. We show how pairwise contrastive methods emerge from learning to produce correct optimal assignments and then extend them to a set level by generalizing an underlying assignment problem, implemented as a regularization for existing methods. As a byproduct of viewing representation learning through the lens of assignment theory, we additionally propose SparceCLR contrastive loss.

Our experiments in instance matching and self-supervised classification suggest that adding quadratic assignment regularization improves the quality of representations learned by backbone methods. We suppose, that our approach would be most useful in the problems where the joint analysis of objects and their groups appears naturally and labeling is not readily available.

\bibliographystyle{splncs04}
\bibliography{egbib}
\newpage
\section*{Supplementary material}

\textbf{Proof of Proposition 1} Let $\textbf{Y}^{*} = \hat{y} (\textbf{S}_m)$ be the solution of the optimal linear assignment problem given the similarity matrix $\textbf{S}_m$. With this, we can rewrite $L(\textbf{S}_m, \textbf{Y}_{gt})$ as follows:

\begin{equation}
L(\textbf{S}_m, \textbf{Y}_{gt}) = \sum_{i} \sum_{j} ([\textbf{Y}_{gt}]_{ij} - [\textbf{Y}^{*}]_{ij}) [\textbf{S}_m]_{ij} = \sum_{i} \sum_{j} q_{ij} [\textbf{S}_m]_{ij}
\end{equation}

Note that the term $q_{ij}$ can only take the values in $\{ -1, 0, 1\}$ as both matrices $\textbf{Y}_{gt}$ and $\textbf{Y}^{*}$ are binary. Consider the case when $q_{ij} = 0$, which indicates that the objects $i$ and $j$ are matched correctly and that $[\textbf{S}]_{ij} + m \leq [\textbf{S}]_{ik}$ for any $k \neq j$. The latter implies that the distance between the positive pair $(i,j)$ is at least $m$ smaller than the distance with all other pairs in the batch.

Next, note that the optimal value of the loss function $L(\textbf{S}_m, \textbf{Y}_{gt}) = 0$ is achieved when all $q_{ij}$ equate to zero, which implies that at the optimum, distances in all positives pairs are at least $m$ smaller than the distances in all negative pairs. This optimality condition reassembles the formulation of margin Triplet loss \cite{hermans2017defense}. $\square$ \\

\noindent \textbf{Proof of Proposition 2} Recall that the batch-hard mining emerges from relaxing constraint $\textbf{Y} \in \Pi$ into $\textbf{Y} \in \mathcal{R}$, where $\mathcal{R}$ is a set of binary row-stochastic matrices. We can rewrite $L_{\tau}(\textbf{S}_m, \textbf{Y}_{gt})$ with batch-hard mining as:

\begin{equation}
\label{eq:proof_prop2}
    \begin{split}
        \hat{L}_{\tau}(\textbf{S}, \textbf{Y}_{gt}) & = \sum_{(p,k) \in \textbf{Y}_{gt}} [\textbf{S}]_{pk} + \tau \sum_{i} log \big{(} \sum_{j} exp(-\frac{1}{\tau} [\textbf{S}]_{ij}) \big{)} \\
        & = \sum_{(p,k) \in \textbf{Y}_{gt}} \phi (p, k) + \tau \sum_{i} log \big{(} \sum_{j} exp(-\frac{1}{\tau} \phi (i, j)) \big{)}
    \end{split}
\end{equation}

where $\phi (p, k)$ is a distance measure between the samples $p$ and $k$, and $\tau$ is the parameter controlling a degree of smoothness. Note that in \eqref{eq:proof_prop2} we assume that the distance $\phi (p, k)$ decreases when the similarity between the samples increases. 

The first and the second terms in \eqref{eq:proof_prop2} relates to the alignment and distribution terms \cite{chen2021intriguing} and reassemble the formulation of the normalized temperature cross entropy loss (InfoNCE) \cite{oord2019representation,chen2020simple} up to a normalization constant. $\square$ \\

\subsection*{Formulations of contrastive losses with batch-hard mining}

Here we present formulations of the contrastive losses with batch-hard mining strategy. We show the derivation for the case of the structured linear assignment loss, while the derivations for the smoothed structured linear assignment and SparseCLR losses can be done analogously. \\

We start from the original structured linear assignment loss with one-to-one mining:

\begin{equation}
    \label{eq:supp_salo}
    L(\textbf{S}, \textbf{Y}_{gt}) = \tr (\textbf{S} \textbf{Y}_{gt}^T) - \underset{\textbf{Y} \in \Pi} {\min} \tr(\textbf{S}\textbf{Y}^T)
\end{equation}

To derive the loss under the batch-hard mining strategy, the structural constraint $\textbf{Y} \in \Pi$ is relaxed to $\textbf{Y} \in \mathcal{R}$, where $\mathcal{R}$ is a set of row-stochastic binary matrices, i.e. $[\textbf{Y}]_{ij} \in \{0,1\}$ for $\forall (i,j)$ and $\sum_{j} [\textbf{Y}]_{ij} = 1$ for $\forall i$. With this, the structured linear assignment loss with the batch-hard mining is defined as follows:

\begin{equation}
\begin{split}
    \label{eq:supp_salo_r}
    \hat{L}(\textbf{S}, \textbf{Y}_{gt}) & = \tr (\textbf{S} \textbf{Y}_{gt}^T) - \underset{\textbf{Y} \in \mathcal{R}} {\min} \tr(\textbf{S}\textbf{Y}^T) \\
    & = \tr (\textbf{S} \textbf{Y}_{gt}^T) - \underset{u_1 ... u_{N}} {\min} \sum_{i} (\sum_{j} [\textbf{S}]_{ij} [u_{i}]_{j}) \\
    & = \tr (\textbf{S} \textbf{Y}_{gt}^T) - \sum_{i} \underset{u_i} {\min} (\sum_{j} [\textbf{S}]_{ij} [u_{i}]_{j}) \\
    & = \tr (\textbf{S} \textbf{Y}_{gt}^T) - \sum_{i} \underset{j} {\min} [\textbf{S}_{ij}]
\end{split}
\end{equation}

where the first inequality follows from the fact that the rows $u_1 ... u_{N}$ of $\textbf{Y} \in \mathcal{R}$ are independent of each other, and the last inequality is due to $u_i$ is a binary vector containing a one-hot encoding of the minimum index. \\

\noindent $\blacktriangleright$ Smoothed structured linear assignment loss $L_{\tau}(\textbf{S}, \textbf{Y}_{gt})$ with batch-hard-mining:

\begin{equation}
    \label{eq:smooth-salo_R}
    \hat{L}_{\tau}(\textbf{S}, \textbf{Y}_{gt}) = \tr(\textbf{S} \textbf{Y}_{gt}^T) + \tau \sum_{i} \log \sum_{j} \exp(- \frac{1}{\tau} [\textbf{S}]_{ij} )
\end{equation}

\noindent $\blacktriangleright$ SparseCLR contrastive loss $L_{\textit{sparse}}(\textbf{S}, \textbf{Y}_{gt})$ with batch-hard mining:

\begin{equation}
    \label{eq:sparseclr_R}
    \hat{L}_{\textit{sparse}}(\textbf{S}, \textbf{Y}_{gt}) = \tr(\textbf{S} \textbf{Y}_{gt}^T) - \frac{1}{2} \sum_{i} \sum_{j \in \Omega(h_i)} \big{(} [h_i]_{j}^2 - \mathcal{T}^{2} (- h_i) \big{)}
\end{equation}

where $h_i \in \mathbb{R}^N$ correspond to a $i$-th row of the similarity matrix $\textbf{S}$, and $\Omega(X) = \{ j \in X : \textit{ sparsemax}(X)_{j} > 0\}$ is the support of \textit{sparsemax} function \cite{Martins2016sparsemax}. The thresholding operator $\mathcal{T}$ is defined as:

\begin{equation}
\label{eq:sparseclr_thresh}
    \mathcal{T}(z) = \frac{(\sum_{j \in \Omega(z)} z_j) - 1}{|\Omega(z)|}
\end{equation}

\noindent And the \textit{sparsemax} function is defined as:

\begin{equation}
\label{eq:sparseclr_sparsemax}
    \textit{sparsemax}(z) = \underset{p \in \Delta^{N-1}} {\argmin} ||p-z||^2
\end{equation}

with $\Delta^{N-1}$ being $N-1$ dimensional simplex.

\section*{QARe for a positive similarity measure}

In the derivations in the main paper, we adopt the metric notation commonly used in assignment problems, where the maximum similarity between objects is indicated by the minimum distance between their representations (e.g. Euclidean distance). In practice, however, the task may require to use of other types of similarity measures, for which the opposite holds (e.g. cosine distance). For this case, the derivation takes the analogous path, but with a switch of signs in Equation \ref{eq:salo} and with $\min$ replaced by $\max$ in the LAP/QAP objectives. With this, the structured quadratic assignment loss equates to:

\begin{equation}
\begin{split}
    \label{eq:lqap_max}
    L_{QAP} & = \underset{\textbf{Y} \in \Pi} {\max} \big{\{} \tr(\textbf{S}\textbf{Y}^T) + \tr(\textbf{S}_{\mathcal{A}} \textbf{Y} \textbf{S}_{\mathcal{B}}^{T} \textbf{Y}^{T}) \big{\}} - \tr(\textbf{S}\textbf{Y}_{gt}) \\
    & \leq \langle \lambda_{\mathcal{A}}, \lambda_{\mathcal{B}} \rangle_{+} + \underbrace{ \underset{\textbf{Y} \in \Pi} {\max} \big{\{} \tr(\textbf{S}\textbf{Y}^T) \big{\}} - \tr(\textbf{S}\textbf{Y}_{gt})}_{\textit{structured linear assignment loss}}
\end{split}
\end{equation}

where $\langle \lambda_{\mathcal{A}}, \lambda_{\mathcal{B}} \rangle_{+}$ corresponds to a maximum dot product between eigenvalues of the matrices $S_{A}$ and $S_{B}$. 

Practically, modifying the QARe computation presented in Algorithm \ref{alg:infonce_qare} in the main paper from the Euclidean distance to the cosine similarity requires 3 steps: (i) similarities are scaled to be non-negative, (ii) computing a maximum dot product instead of a minimum dot product, (iii) the maximum dot product of eigenvalues is added to a contrastive objective. Note that step (i) is extra compared to the Euclidean distance case and is needed because the QAP formulation requires non-negative distances as an input. The algorithm is summarized in Algorithm \ref{alg:infonce_qare_cosine}.

\section*{Experiments}

\subsection*{Time and memory complexity}

\begin{figure}[h!]
  \centering
    \includegraphics[width=0.5\linewidth]{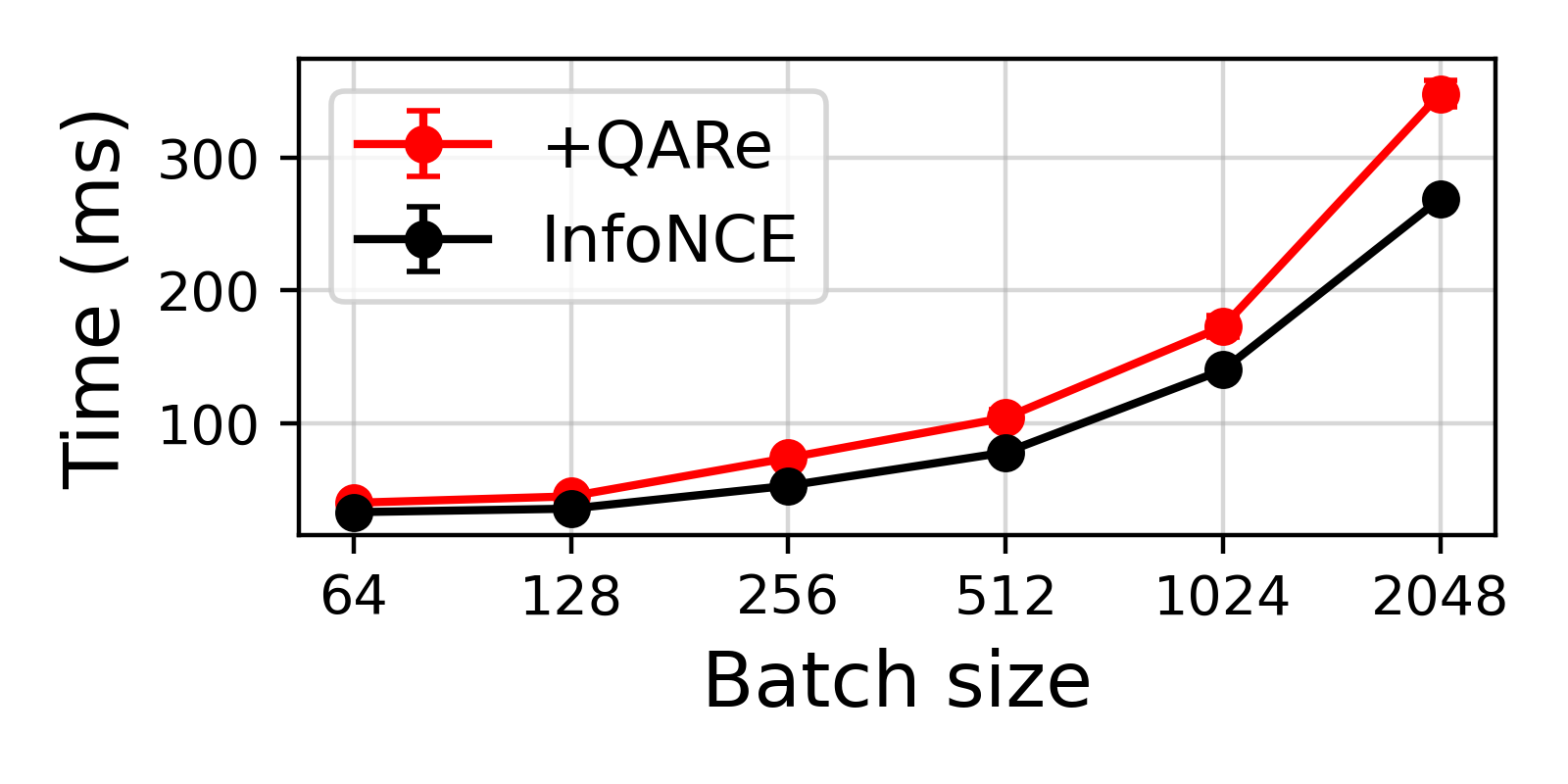}
    \caption{A training iteration time complexity of InfoNCE with and without the quadratic assignment regularization.}
  \label{fig:time}
\end{figure}

In addition to computational complexity, we provide an empirical analysis of how the proposed regularization influences the time and memory complexity of a baseline method (InfoNCE). We observe $13\%$ increase in training time for the batch size of $256$, and up to $29\%$ for the batch size of $2048$ (Figure \ref{fig:time}). The increase in memory consumption is negligible ($+0.78\%$ for the batch size of $2048$).

\subsection*{Instance matching}
\paragraph{Augmentations.} During the training, we apply a simple augmentation strategy: horizontal flip with a 50\% chance and color jittering. For the latter, the brightness, contrast, saturation, and hue are sampled from a uniform distribution $U[0, 0.1]$.

\paragraph{Hyper-parameters of the losses.} We use 4 baseline contrastive losses: margin Triplet \cite{hermans2017defense}, InfoNCE \cite{oord2019representation}, NT-Logistic \cite{mikolov2013efficient} and the proposed SparseCLR losses, which we extend with the Quadratic Assignment Regularization (QARe). The margin parameter for Triplet loss is set to 0.5, the temperature parameters for InfoNCE and NT-Logistic are both set to 0.05. We combine QARe and the backbone contrastive learning loss by taking their convex combination, where the QARe term is weighted by the constant $\beta \geq 0$. The exact values of $\beta$ are $0.4 / 0.5 / 0.2 / 0.3$ for margin Triplet / InfoNCE / NT-Logistic / SparseCLR.

\begin{table*}[b]
 \begin{adjustbox}{width=1\columnwidth,center}
  \begin{tabular}{lcccccc}
    \toprule
     & \multicolumn{3}{c}{\textbf{Conv-4}} & \multicolumn{3}{c}{\textbf{ResNet-32}} \\
    \cmidrule[0.1pt](r){2-4} \cmidrule[0.1pt](l){5-7}
    \textbf{Method} &
    \textbf{CIFAR-10} & \textbf{CIFAR-100} & \textbf{tiny-ImageNet} &
    \textbf{CIFAR-10} & \textbf{CIFAR-100} & \textbf{tiny-ImageNet} \\
    \midrule
    
    ($\beta$) SimCLR+QARe & 
    0.5 & 
    0.375 & 
    0.875 &
    1.125 & 
    1.125 & 
    1.125 
    \\
    \cmidrule(l){1-7}
    ($\beta$) SparseCLR+QARe & 
    0.125 & 
    1.25 & 
    0.5 &
    1 & 
    0.875 & 
    1.25 
    \\
    \bottomrule
  \end{tabular}
 \end{adjustbox}
 \centering
 \caption{QARe weighting for the architectures and the datasets for self-supervised classification experiment.}
\label{tab:qare_beta}
\end{table*}

\subsection*{Self-supervised classification}

\paragraph{Encoder architectures.} We use two models: Conv-4 and ResNet-32. The Conv-4 model involves 4 blocks. The first 3 blocks consist of 8, 16, and 32 feature maps respectively. Each performs a convolution with a kernel size of 3, a stride of 1, and a padding of 1 pixel, followed by a batch-normalization layer, a ReLU, and an average pooling layer with a kernel size of 2 and a stride of 2 pixels. The fourth block performs the same operations but instead of an average pooling, an adaptive average pooling is used. For ResNet-32, we use off-the-shelf Pytorch implementation as described in \cite{He2015resnet}. We initialize the model using standard Xavier initialization \cite{glorot2010understanding} and set batch-normalization weights and biases to 1 and 0 respectively.

\paragraph{Augmentations.} For self-supervised training we apply the following augmentations to images: horizontal flip with a 50\% chance, random crop-resize, grayscale conversion with a 20\% chance, and color jitter with an 80\% chance. Random crop-resize consists of cropping the given image from 0.08 to 1.0 of the original size, then changing its aspect ratio from 3/4 to 4/3 of the original, and finally resizing to input shape using a bilinear interpolation. For color jitter, the brightness, contrast and saturation are sampled from $U[0, 0.8]$, and the hue is sampled from $U[0, 0.2]$.

\paragraph{Hyper-parameters of the losses.}

We combine quadratic assignment regularization and the backbone contrastive learning loss by summing with weighting the QARe term by a constant $\beta \geq 0$. The values of $\beta$ for each architecture and dataset are listed in Table \ref{tab:qare_beta}. The temperature parameter for both SimCLR \cite{chen2020big} and SimCLR+QARe is set to 0.05.

\begin{algorithm}[h]
   \caption{Pseudocode for set-based InfoNCE with cosine similarity and Quadratic Assignment Regularization (QARe).}
   \label{alg:infonce_qare_cosine}
   
    \definecolor{codeblue}{rgb}{0.25,0.5,0.5}
    \lstset{
      basicstyle=\fontsize{7.2pt}{7.2pt}\ttfamily\bfseries,
      commentstyle=\fontsize{7.2pt}{7.2pt}\color{codeblue},
      keywordstyle=\fontsize{7.2pt}{7.2pt},
    }
\begin{lstlisting}[language=python]
# f: encoder network
# alpha: weighting for the pairwise contrastive part (linear assignment)
# beta: weighting for the set contrastive part
# N: batch size
# E: dimensionality of the embeddings

for x in loader: # load a batch with N samples
    # two randomly augmented views of x
    y_a, y_b = augment(x)
    
    # compute embeddings
    z_a = f(y_a) # NxE
    z_b = f(y_b) # NxE
    
    # compute inter-set and intra-set similarities
    S_AB = similarity(z_a, z_b) # NxN
    S_A  = similarity(z_a, z_a) # NxN
    S_B  = similarity(z_b, z_b) # NxN
    
    # compute pairwise contrastive InfoNCE loss
    pairwise_term = infonce(S_AB)
    
    # shift to non-negative & compute eigenvalues
    eigs_a = eigenvalues(1 + S_A) #N
    eigs_b = eigenvalues(1 + S_B) #N
    
    # compute QARe from maximum dot product of eigenvalues
    eigs_a_sorted = sort(eigs_a, descending = True) #N
    eigs_b_sorted = sort(eigs_b, descending = True) #N
    qare = eigs_a_sorted.T@eigs_b_sorted
    
    # combine pairwise contrastive loss with QARe
    loss = alpha*pairwise_term + beta*qare/(N^2)
    
    # optimization step
    loss.backward()
    optimizer.step()
\end{lstlisting}
\end{algorithm}

\end{document}